# ProGAN-Based Synthetic Image Generation for Industrial Visual Inspection of Metallic Parts

Hossein Omid Beiki [a]*, Sepehr Asgari [a], Oliver Petrovic [a], Christian Brecher [a]

[a] *Laboratory for Machine Tools and Production Engineering WZL of RWTH, Aachen University, Aachen, Germany*

* Corresponding author. Tel.: +49-241-8027449; *E-mail address:* h.omid@wzl.rwth-aachen.de

**Abstract**

Automated visual inspection is essential for efficient repair and recycling of metallic components. AI-based defect detection is a strong fit; however, it is often constrained by scarce, low-diversity training data. To address this gap, we study Generative AI–driven synthetic image generation using Progressive Growing GANs (ProGAN). We implement a ProGAN pipeline with Wasserstein loss and gradient penalty and evaluate key hyperparameters on two public datasets, MVTec AD (metal nut) and MPDD (tube). As a result, training remained stable and the model yielded realistic images. These images can augment training data, improve robustness, and reduce the cost of collecting rare defects.

## 1. Introduction

One of the major bottlenecks in recycling and remanufacturing processes lies in the inspection of components with varying shapes, materials, and conditions. Traditionally, inspection tasks have been performed by trained human operators. While human vision is highly adaptable, manual inspection is limited by fatigue, subjectivity, inconsistent performance [1,2].

To address these limitations, automated visual inspection systems have been increasingly adopted in industrial quality assurance. Recent advances in Artificial Intelligence (AI) have enabled automated inspection systems to analyze complex visual data and detect subtle variations that may indicate defective components. However, such systems are fundamentally data-driven and require large amounts of representative training data to achieve reliable performance.

In many industrial scenarios, collecting sufficient defect data is difficult because defects occur rarely. As a result, only a limited number of defect samples are typically available, while the collection and annotation of defect images require significant time, labor, and domain expertise [3,4]. Consequently, data scarcity remains one of the central challenges in applying AI-based visual inspection systems in real-world production scenarios.

Several approaches exist to address data scarcity through data augmentation. Traditional methods rely on transformation-based techniques that modify existing images, while data synthesis approaches generate new samples, for example through graphics-based simulation or generative AI models (Fig. 1). Among synthesis-based approaches, Generative Artificial Intelligence (GenAI) models provide a way to generate synthetic data by learning statistical distribution of the observed dataset [5]. These models can produce new samples that follow similar patterns to the original data [6]. Among these models, Generative Adversarial Networks (GANs) have shown strong capabilities in producing realistic synthetic images [7].

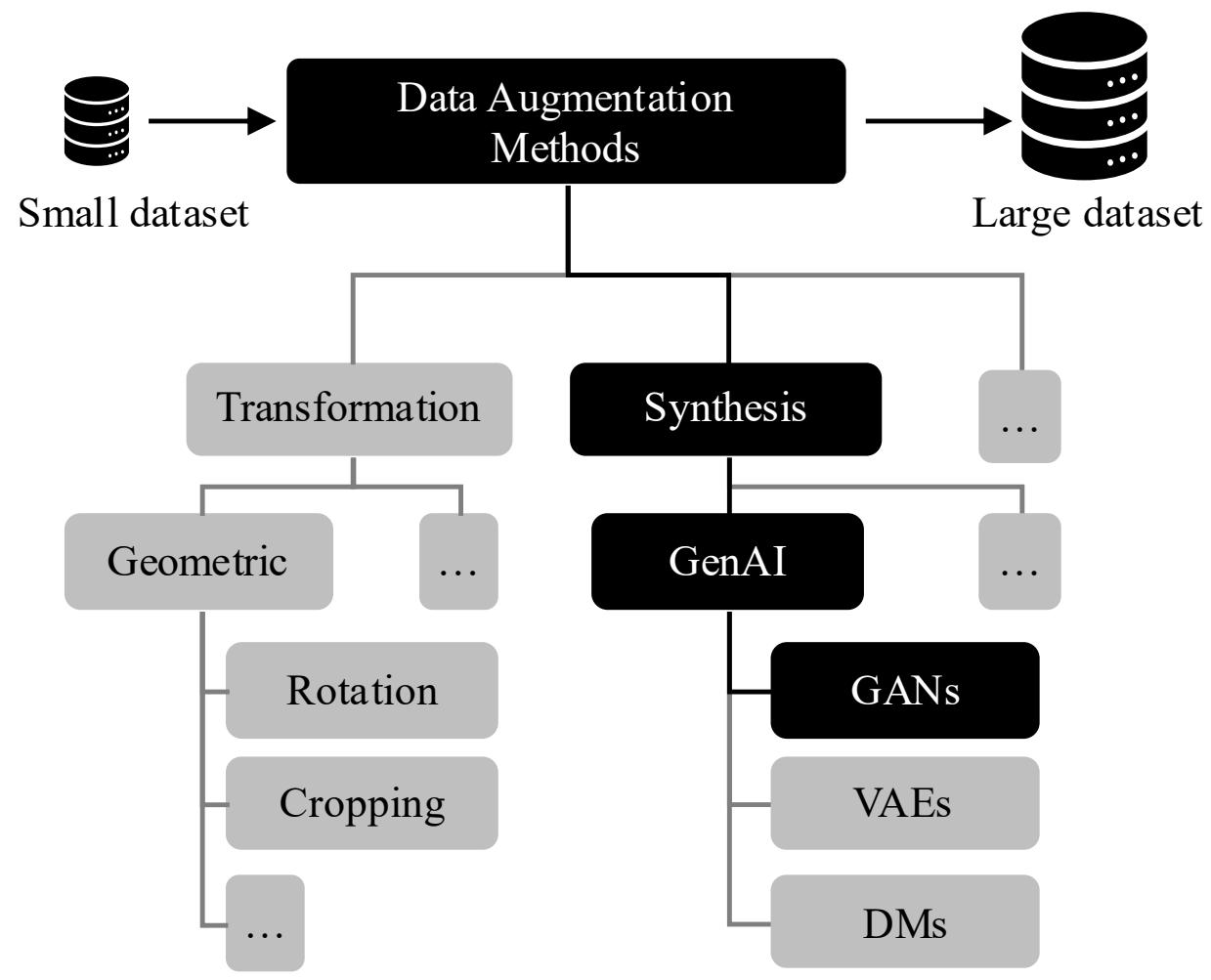

Fig. 1 Overview of different data augmentation and synthetic data generation methods [5]

A GAN consists of two neural networks—a generator and a discriminator—trained in an adversarial process in which the generator learns to produce samples that resemble real data. However, conventional GAN models often suffer from training instabilities such as convergence failure and mode collapse. Progressive Growing GAN (ProGAN), introduced by Karras et al. [8], addresses these issues by progressively increasing the image resolution during training. This approach allows the model to first learn coarse image structures before capturing finer visual details. Compared to alternative generative models such as Variational Autoencoders or diffusion models, ProGAN provides a suitable balance between image quality, training stability, and computational efficiency for scenarios with limited datasets [9].

Despite increasing research on synthetic data generation, only limited work has investigated the use of ProGAN in industrial inspection scenarios, particularly for metallic components under scarce-data conditions. To address this gap, this paper examines the feasibility of ProGAN-based synthetic image generation for industrial visual inspection.

The contributions of this work are summarized as follows. First, a ProGAN-based synthetic image generation pipeline for industrial visual inspection under scarce-data conditions is designed and implemented using Wasserstein loss with gradient penalty. Second, the feasibility of the approach is evaluated on two public datasets with different levels of visual complexity, namely MVTec Anomaly Detection (MVTec AD) [10] and Mtal Parts Defect Detection (MPDD) [11]. Third, the usefulness of the generated images is assessed through qualitative inspection, Kernel Inception Distance (KID) [12], and a downstream visual inspection task based on YOLO.

This work focuses on synthetic image generation only; annotation generation is left for future work.

The remainder of this paper is organized as follows. Section 2 reviews related work, Section 3 describes the methodology, Section 4 presents the results, and Section 5 concludes the paper.

## 2. Related Work

Generative models have increasingly been investigated to address data scarcity in industrial visual inspection. GAN-based approaches have shown strong potential for generating synthetic defect samples that complement limited real datasets. One representative example is the work of Mătase and Andonie [13], who investigate synthetic data generation for visual metal defect detection under limited data conditions. They use a GAN-based approach to generate synthetic defect images together with corresponding pixel-level masks and combine these data with additional augmentation in the training pipeline of a defect detection model. Their results show improvements in both image-level and pixel-level performance, indicating that synthetic defect generation can effectively support industrial inspection when real defect samples are scarce.

Several other studies also report promising results for GAN-based synthetic defect generation. Jain et al. [14] investigated a GAN-based augmentation pipeline for surface defect detection and classification on the NEU-CLS steel defect dataset. Their study trained multiple GAN architectures to synthesize additional defect samples and showed that the generated images can support downstream classification when real data are limited.

Xia et al. [15] addressed railway surface defect detection with RailGAN, a model derived from CycleGAN that introduces a pre-sampling stage to place synthetic defects more consistently on the rail surface and avoid unrealistic background artifacts.

Wang et al. [16] proposed an improved GAN (iGAN) for machined surface inspection, combining adversarial learning with positive-sample-based image restoration to support defect detection under scarce defective data.

Rippel et al. [17] focused on anomaly detection in fabrics and used a guided image-to-image translation approach to transfer defect appearance across different fabric types, showing that synthetic anomalies can improve downstream inspection robustness.

Du et al. [18] introduced Con-GAN for surface defect recognition under limited data, combining shared data augmentation with additional regularization to reduce overfitting and improve the diversity and usefulness of generated defect samples.

Overall, these studies show that GAN-based synthetic data generation is a promising strategy for addressing limited training data in industrial inspection. However, only limited work has evaluated Progressive Growing GANs (ProGAN) for metallic component inspection under scarce-data conditions. This motivates the present study, which investigates the feasibility of ProGAN-based synthetic image generation for industrial visual inspection of metallic parts.

## 3. Methodology

This section describes the proposed methodology for synthetic image generation based on ProGAN. An overview of the methodology is shown in Fig. 2. The individual components, including the model architecture, the training design under scarce-data conditions, the datasets used in the experiments, and the evaluation procedure, are described in the following subsections. The objective is to answer two central questions. First, can ProGAN be trained in a stable manner on small industrial datasets? Second, are the generated images realistic enough to be useful for data augmentation in a practical visual inspection scenario?

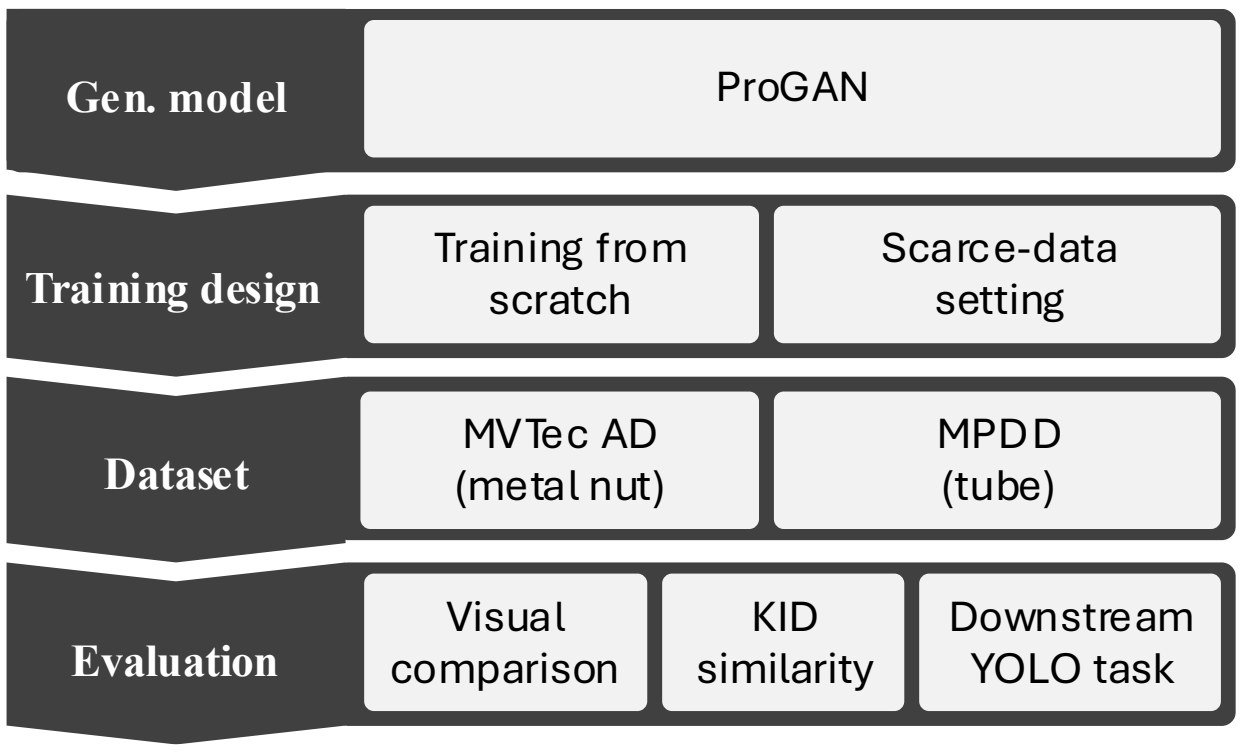

Fig. 2 Overview of the proposed methodology for ProGAN-based synthetic image generation and evaluation

### 3.1. ProGAN model and training objective

ProGAN was selected as the main generative model due to their improved training stability and its ability to generate high-resolution images from limited data. In contrast to conventional GAN architectures, ProGAN progressively increases the resolution of generated images during training (Fig. 3). Training starts at a low spatial resolution, and additional layers are gradually added to both the generator and discriminator. This progressive learning strategy allows the model to first learn global image structures before focusing on finer visual details. As a result, training becomes more stable and typical GAN problems such as convergence failure and mode collapse can be reduced.

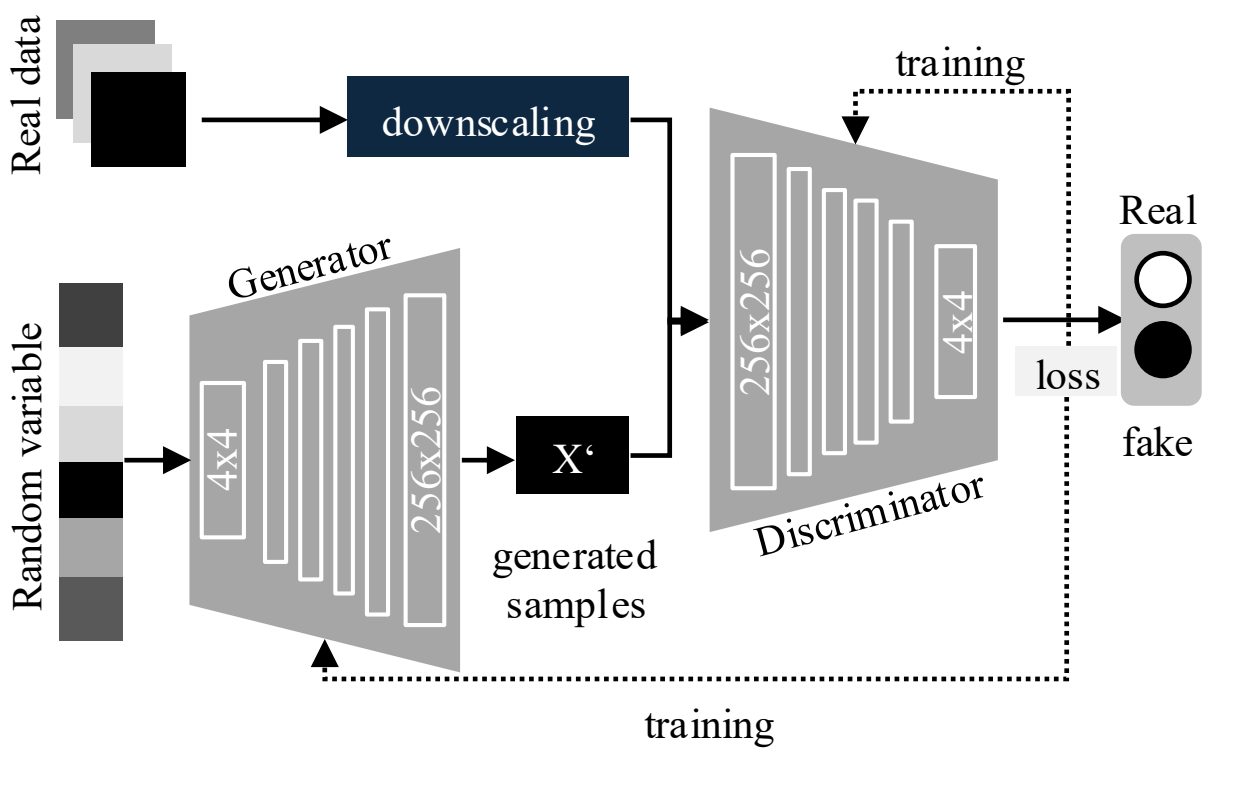

Fig. 3 Illustration of the progressive growing strategy in ProGAN training [8]

To further improve training stability, the model was trained using Wasserstein loss with gradient penalty (WGAN-GP). Traditional GAN loss functions often lead to unstable optimization and vanishing gradients. Wasserstein loss addresses this issue by approximating the Wasserstein distance between real and generated data distributions. The gradient penalty term enforces the Lipschitz constraint required for the Wasserstein formulation. This combination improves gradient behaviour and helps stabilize GAN training when only limited training data is available.

### 3.2. Training design and expected challenges

This work focuses on synthetic image generation under scarce-data conditions, where the limited number of available samples makes GAN training particularly difficult. In such scenarios, the discriminator can quickly memorize the dataset, which may lead to unstable training and poor generalization of the generator. To obtain a stable training configuration, a set of training parameters was adjusted during the experiments, with particular attention to learning rate and resolution. Data augmentation was used cautiously, since stronger augmentation may introduce visible artifacts into generated samples. The final configuration was selected based on training stability, visual quality, and hardware feasibility.

Different training strategies were considered, including training from scratch and fine-tuning. However, no relevant pretrained ProGAN models for industrial defect imagery were publicly available. Therefore, the reported experiments were conducted using training from scratch, which proved to be the most reliable and reproducible approach for the selected datasets.

### 3.3. Datasets

Two public datasets with metallic parts were used for evaluation: (1) the metal nut category from MVTec AD and (2) the tube category from MPDD. These two datasets were selected because they represent different levels of visual complexity and therefore allow evaluation under both controlled and more challenging conditions.

The MVTec AD metal nut category contains images of a single object on a relatively uniform background. As a result, the visual complexity is comparatively low, which makes this dataset suitable as a more controlled scenario. In contrast, the MPDD tube category contains larger variations in background, lighting conditions, and object positioning. This makes MPDD a more challenging dataset and therefore useful for examining the robustness of the proposed model under less controlled conditions.

For MVTec AD metal nut, the training split contains 220 good images, while the test split contains 22 good and 93 anomalous images at a resolution of 700 × 700 pixels. For MPDD tube, the training split contains 122 good images, and the test split contains 32 good and 69 anomalous images at a resolution of 1024 × 1024 pixels. Since the objective of the generative model in this study is to learn the visual distribution of the available component appearances rather than to optimize a discriminative classifier directly, all available images were considered during generative data preparation to improve coverage of the limited visual manifold. The downstream YOLO-based evaluation, however, was analyzed separately (see 3.4).

Before training, the images were resized to the target resolution used in the experiments. The final selected output resolution was 256 × 256 pixels, which provided a practical balance between visual applicability and the memory limitations of the available hardware.

### 3.4. Evaluation protocol

The generated images were evaluated using both qualitative and quantitative methods, followed by an application-oriented downstream experiment. This three-level evaluation was chosen to assess the generated samples not only visually, but also statistically and practically.

First, a qualitative evaluation was performed through visual inspection of generated samples. The purpose of this step was to analyze whether the generator could reproduce the overall visual structure and characteristic appearance of the real industrial components. This includes aspects such as object shape, relative positioning of visible structures, and the general plausibility of the generated samples.

Second, quantitative evaluation was conducted using Kernel Inception Distance (KID). KID measures the distance between the feature distributions of real and generated images. In contrast to the widely used Fréchet Inception Distance (FID), KID provides an unbiased estimator with respect to dataset size, which makes it more suitable for small datasets such as those used in this study. Lower KID values indicate a smaller discrepancy between the feature distributions of real and generated samples and therefore a higher similarity between both sets.

Third, the practical usefulness of the generated synthetic images was evaluated in a downstream visual inspection task. For this purpose, two YOLO-based classification setups were considered: a baseline model trained only on real data and an augmented model trained on real data together with synthetic images generated by ProGAN. Ultralytics YOLO11n-cls [19] was selected because the considered MVTec task is a classification task and YOLO-based models are widely used in industrial computer vision applications [20]. Both models were trained under identical conditions to ensure a fair comparison. The objective of this experiment was to investigate whether the inclusion of synthetic images can improve the behavior of a practical inspection model.

### 3.5. Experimental Setup

The ProGAN model was trained using progressive resolution growth, starting from low resolutions and gradually increasing to higher resolutions. Depending on the experiment, common augmentation strategies such as cropping and rotation were applied to increase dataset diversity. The best-performing configuration on MVTec AD used an image resolution of 256 × 256 pixels. Since the hardware resources were limited, the final configuration was selected not only based on image quality but also based on computational feasibility.

For the downstream YOLO models, the selected hyperparameters were 50 epochs, an image size of 120, and a batch size of 2, while the remaining parameters were kept at their default values.

All experiments were conducted on a workstation with an Intel Core i7-12700K processor and an NVIDIA RTX 3060 GPU with 12 GB of VRAM. The available hardware influenced several design choices, including the final image resolution and the number of training epochs. The software environment consisted of Python 3.11.9, PyTorch 2.6.0, and CUDA 11.8.

## 4. Results and Discussion

This section presents the experimental results and discusses them with respect to training stability, visual quality, quantitative similarity, and practical usefulness in a downstream inspection task. Two ProGAN models were trained on MVTec AD and MPDD. Overall, the proposed pipeline showed a stable training behavior under the considered settings and was able to generate visually meaningful synthetic images in both cases.

### 4.1. Training stability

A first important observation is that stable training could be achieved on both datasets despite the scarce-data setting. This is a relevant result, because small industrial datasets are especially prone to unstable GAN optimization. The progressive training strategy combined with WGAN-GP contributed to a more controlled training behavior and allowed the model to learn meaningful structures without immediate collapse.

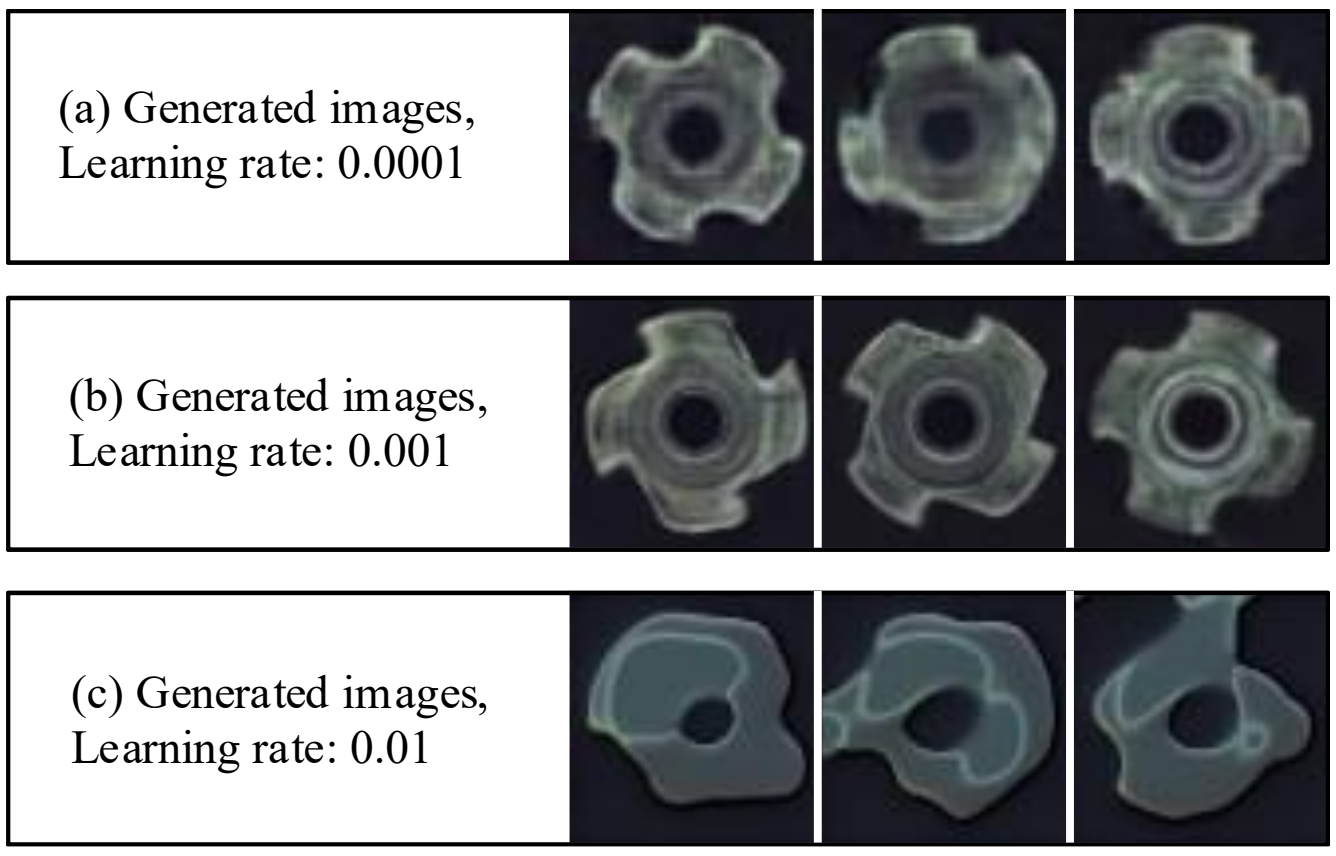

Fig. 4: effect of learning rate on the generated images

Among all investigated hyperparameters, the learning rate had the strongest effect on the training process. Small changes in the learning rate could substantially influence whether the generator learned meaningful visual structures or produced poor-quality outputs. Fig. 4 shows that unsuitable learning rates led to weak image formation, whereas an appropriate learning rate enabled the model to capture the basic appearance of the object already at lower resolution. This observation confirms that hyperparameter tuning remains critical when ProGAN is applied to small industrial datasets.

### 4.2. Qualitative comparison between real and generated images

For MVTec AD metal nut, the generated images show that the overall shape of the metal nuts was learned well (Fig. 5). In particular, the model was able to reproduce the global structure of the object and the general arrangement of the visible component regions. Even when trained on a relatively small dataset, the generator learned visually plausible samples that follow the dominant structure of the original data distribution. This indicates that ProGAN can capture coarse structural information reliably in controlled industrial image settings

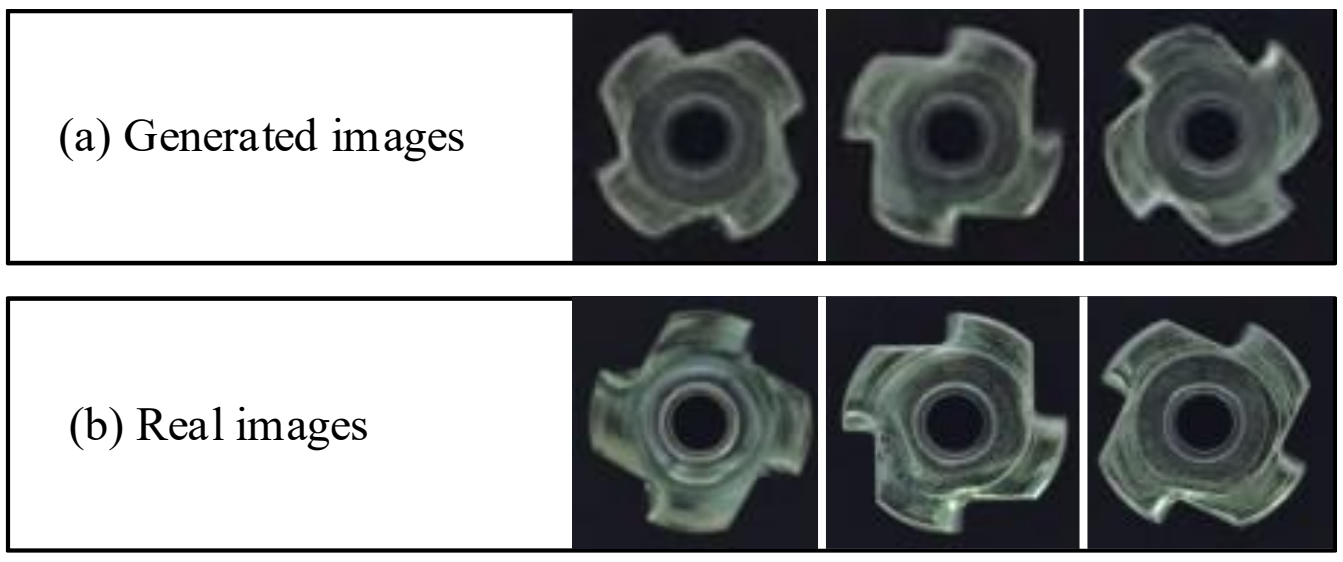

Fig. 5 generated vs. real images based on MVTec metal nut dataset (resolution 64×64)

The results on MPDD show a similar overall tendency, although the dataset is more complex. In this case, the model was able to generate samples with some degree of variation, for example by changing the orientation of visible parts or by omitting certain object elements (Fig. 6). However, the generated diversity remained limited.

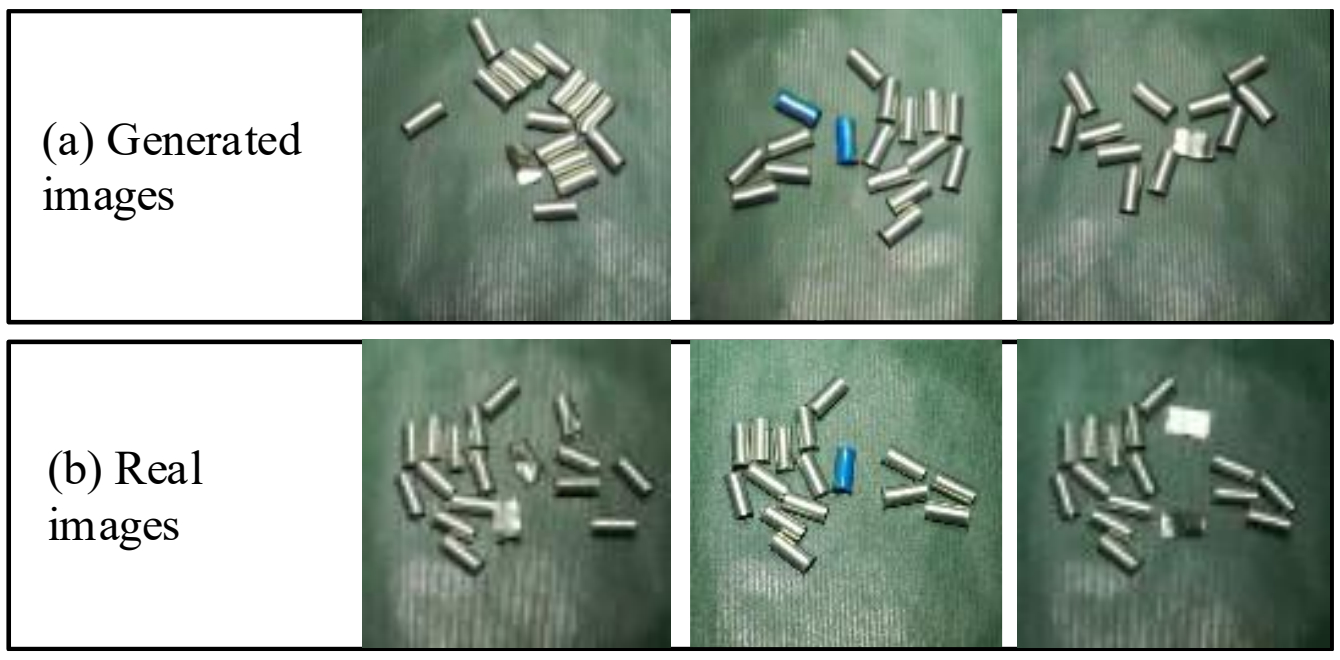

Fig. 6 Difference between a real image (a) and a generated (b) image by the proposed ProGAN model

A limitation observed during the experiments was that stronger augmentation in the training process could introduce visible artifacts into some generated samples, which is called Augmentation leakage in the literature. However, this effect is only briefly noted here due to the limited scope of the paper.

*4.3. Quanitative comparison of real vs generated images*

Quantitative evaluation was performed on the MVTec AD metal nut dataset by comparing real images with synthetic samples generated by the trained ProGAN model. Kernel Inception Distance (KID) was used to measure the similarity between real and generated image distributions. The mean KID was 0.125 with a standard deviation of 0.0102. This indicates that the generated samples do not fully match the real data distribution but still show a reasonably close similarity in feature space. Given the limited dataset size and the challenging scarce-data setting, this result can be considered promising.

In addition to image similarity, the practical efficiency of the generation process was evaluated. After training, the model required only 7.49 s to generate 500 images. This shows that, once trained, the generator can produce additional samples with low computational effort.

Overall, these results suggest that the proposed ProGAN pipeline provides a useful approximation of the real image distribution while remaining efficient enough for practical synthetic data generation in industrial inspection scenarios.

*4.4. Evaluation based on downstream visual inspection task*

To further assess the practical usefulness of the generated images, a downstream experiment was conducted on the MVTec AD metal nut dataset using a YOLO-based classification model. Two training configurations were compared: a baseline model trained on real images only and an augmented model trained on the same real images together with synthetic samples generated by ProGAN. Evaluation was performed on the MVTec test set.

Fig. 7 shows the Top-1 accuracy during training for both configurations. The model trained with augmented data exhibited a somewhat more stable training behavior than the baseline model, while both models reached a comparable final Top-1 accuracy. This suggests that the synthetic images mainly replicate the statistical properties of the original dataset rather than introducing substantial new variability. At the same time, the fact that the augmented data does not degrade performance indicates that the generated images follow realistic visual patterns and are consistent with the original data distribution.

Overall, the results indicate that stronger downstream improvements will likely require additional diversity in the generated data. One possible direction is domain randomization, for example by varying backgrounds or applying controlled perturbations of visual properties.

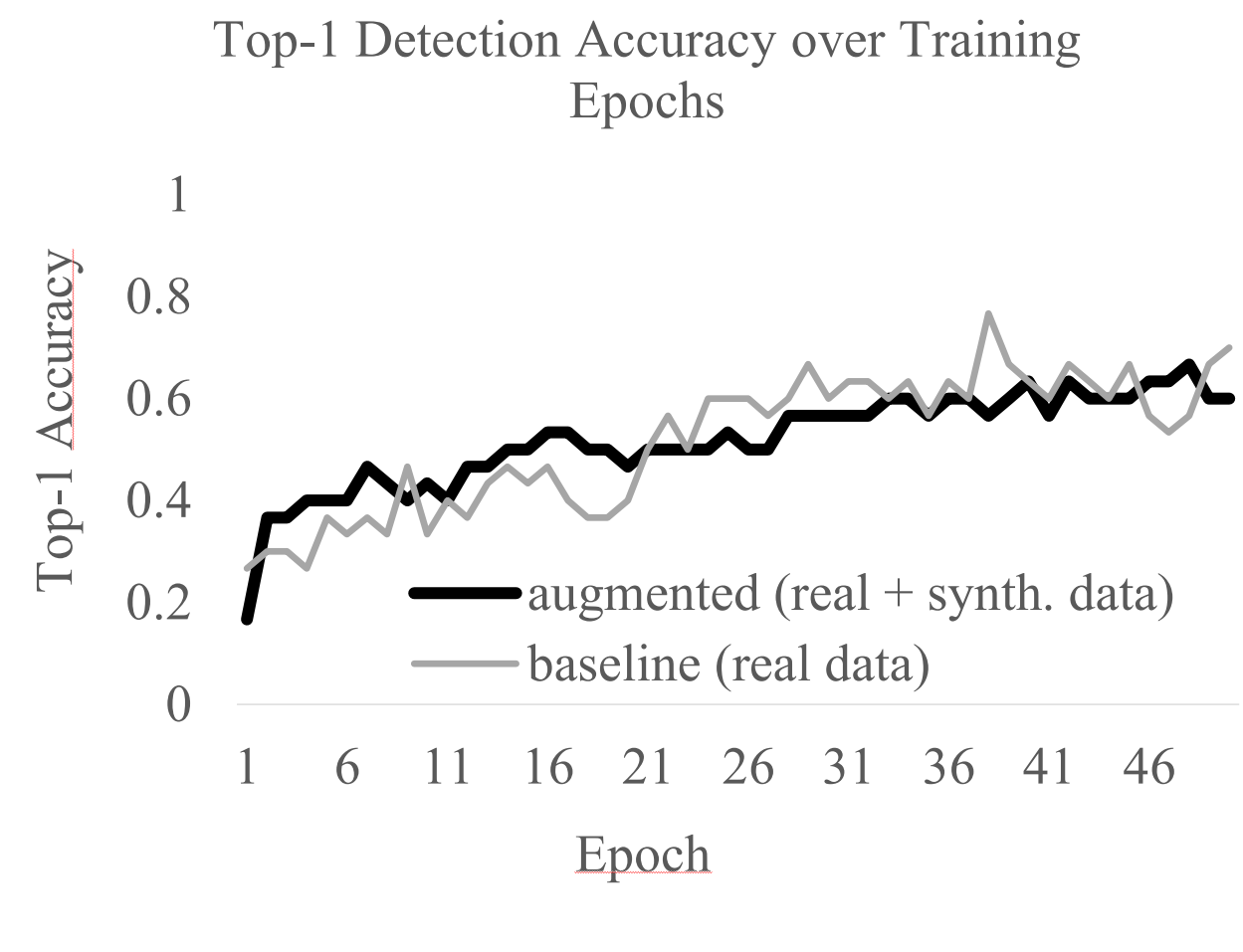

Fig. 7 Top-1 accuracy of the YOLO-based classification model during training on the MVTec AD metal nut dataset for the baseline and augmented setups.

## 5. Conclusion

This work investigated the use of Progressive Growing Generative Adversarial Networks (ProGAN) for synthetic image generation in AI-based visual inspection under scarce-data conditions. A ProGAN-based pipeline using Wasserstein loss with gradient penalty was implemented and evaluated on the MVTec AD and MPDD datasets. The results show that stable training and visually plausible image synthesis are feasible at a resolution of 256 × 256, even when only limited industrial training data are available.

The qualitative and quantitative results indicate that the generated images capture the dominant visual structure of the inspected components while remaining reasonably close to the real data distribution. The achieved KID value of 0.125 ± 0.0102 and the generation time of 7.49 s for 500 images further show that the proposed approach is not only promising in terms of image similarity, but also practical with respect to computational efficiency.

The downstream evaluation on the MVTec AD metal nut task showed that synthetic images can be used for data augmentation without degrading downstream performance. At the same time, the comparable final accuracy of the baseline and augmented models suggests that the generated samples mainly reinforce the existing data distribution rather than introducing substantial new variability.

Overall, the study shows that ProGAN is a feasible approach for synthetic image generation in industrial visual inspection when defect data are scarce. Its main strength lies in generating realistic additional samples from a limited visual distribution, while the diversity of the generated images is still limited by the amount of available training data. Future work may therefore investigate larger datasets and additional diversity mechanisms, such as domain randomization or hybrid approaches combining GAN-based synthesis with foundation models. Another relevant direction is the generation of annotated synthetic data, for example through class-specific or conditional generative models that enable more controllable defect sample generation.

## Acknowledgements

The IGF-projekt 01IF23532N (HyPairBot) of the research association FVP is funded via the DLR by the Federal Ministry for Economic Affairs and Energy (BMWE) due to a decision of the German Bundestag.

.